\pdfoutput=1
\documentclass[letterpaper, 10 pt, journal, twoside]{ieeetran}
\IEEEoverridecommandlockouts                              

\usepackage[english]{babel}
\usepackage[T1]{fontenc}

\usepackage[nolist]{acronym}
\usepackage[cmex10]{amsmath}
\usepackage{amssymb}
\usepackage{amsmath}
\usepackage{bm}
\usepackage{booktabs}
\usepackage{color}
\usepackage[font={small}]{caption}
\usepackage{subcaption}
\usepackage{float}
\usepackage{graphicx}
\usepackage{hyperref}
\usepackage{import}
\usepackage{mathtools}
\usepackage{multirow}
\usepackage{outline}
\usepackage{paralist}
\usepackage{siunitx}
\usepackage{stmaryrd}
\usepackage{systeme}
\usepackage{url}
\usepackage{tikz}
\usepackage{booktabs}
\usepackage{cite}
\usetikzlibrary{tikzmark}
\ifCLASSINFOpdf
\else
\fi

\title{Creating a Dynamic Quadrupedal Robotic Goalkeeper with Reinforcement Learning}%
\author{Xiaoyu Huang$^{2*}$, Zhongyu Li$^{1*}$, Yanzhen Xiang$^{1}$, Yiming Ni$^{1}$, Yufeng Chi$^{1}$, Yunhao Li$^{1}$, Lizhi Yang$^{1}$, \\Xue Bin Peng$^{3}$, and Koushil Sreenath$^{1}$
\thanks{This work was in part supported by NSF Grant CMMI-1944722.}%
\thanks{$^*$ Authors contributed equally}
\thanks{$^1$ University of California, Berkeley, $^2$ Georgia Institute of Technology, $^3$ Simon Fraser University. zhongyu\_li@berkeley.edu.}}

\begin{document}
\maketitle

\begin{abstract}
We present a reinforcement learning (RL) framework that enables quadrupedal robots to perform soccer goalkeeping tasks in the real world. 
Soccer goalkeeping using quadrupeds is a challenging problem, that combines highly dynamic locomotion with precise and fast non-prehensile object (ball) manipulation.
The robot needs to react to and intercept a potentially flying ball using dynamic locomotion maneuvers in a very short amount of time, usually less than one second. 
In this paper, we propose to address this problem using a hierarchical model-free RL framework. 
The first component of the framework contains multiple control policies for distinct locomotion skills, which can be used to cover different regions of the goal.
Each control policy enables the robot to track random parametric end-effector trajectories while performing one specific locomotion skill, such as jump, dive, and sidestep.
These skills are then utilized by the second part of the framework which is a high-level planner to determine a desired skill and end-effector trajectory in order to intercept a ball flying to different regions of the goal. 
We deploy the proposed framework on a Mini Cheetah quadrupedal robot and demonstrate the effectiveness of our framework for various agile interceptions of a fast-moving ball in the real world.
\end{abstract}

\section{Introduction}

Developing a robotic goalkeeper is an appealing but challenging problem. This task requires the robot to perform highly agile maneuvers such as jumps and dives in order to accurately intercept a fast moving ball in a short amount of time.
Solving this problem is attractive because it can offer us solutions to combine dynamic legged locomotion with fast and precise non-prehensile arm manipulation.
Recent developments in quadrupedal robots, which allow for more agile and versatile maneuvers, provides a suitable hardware platform for tackling this task.
Furthermore, recent advances in model-free reinforcement learning~(RL) has shown promising results on developing controllers for dynamic motor skills on quadrupedal robots~\cite{peng2020learning, margolisyang2022rapid, ji2022concurrent}.
However, previous efforts on applying RL on quadrupedal robots mainly focus on low-level locomotion control, such as tracking a desired walking velocity~\cite{ji2022concurrent} or mimicking a reference motion~\cite{peng2020learning}, without extending the learned locomotion skills to a higher level task, such as precisely intercepting a fast-moving soccer ball using agile maneuvers. 
This is challenging because it is a combination of highly dynamic locomotion control and accurate non-prehensile manipulation of a fast moving object, each of which is already a difficult task on its own.  
Therefore, there have been few prior attempts on developing goalkeeping controllers with agile maneuvers using quadrupeds in the real world. 

In this work, we propose to address the goalkeeping task using a hierarchical model-free RL framework. 
This framework decomposes the goalkeeping task into two sub-problems: 1) low-level locomotion control to enable the robot to perform various agile and highly-dynamic locomotion skills, and 2) high-level planning to decide an optimal skill and motion to perform in order to intercept the ball.

\begin{figure}
    \centering
    \includegraphics[width=0.7\linewidth]{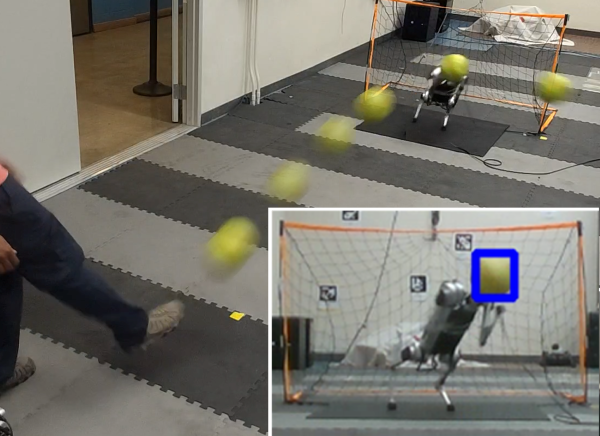}
    \caption{A quadrupedal robot goalkeeper, Mini Cheetah, saves a flying soccer ball towards the goal using the proposed hierarchical RL framework with multiple locomotion control policies and a motion planning policy. The ball flying time is only around $0.5$ second. Video is at \url{https://youtu.be/iX6OgG67-ZQ}.}
    \vspace{-0.4cm}
    \label{fig:intro}
\end{figure}

\subsection{Related Work}

The soccer goalkeeping problem using quadrupedal robots can be viewed as a combination of three domains of robotics research: robotic manipulation to intercept a fast moving object, locomotion control to enable a quadruped to perform highly dynamic maneuvers, and the robot soccer.

\subsubsection{Robotic Catching/Hitting of Fast Moving Objects}
Enabling robots to catch or hit fast moving objects, such as a ball, has been studied extensively in the robotic manipulation field. Typically, robotic arms, with a fixed base~\cite{gao2021optimal} or a mobile base~\cite{9341134}, and quadrocopter~\cite{ritz2012cooperative} are used for these tasks. A common approach to tackling catching tasks is to separate it into two sub-tasks: prediction of the ball's trajectory based on the estimated ball position and velocity using models of the ball's dynamics~\cite{kocc2018online, gao2021optimal, huang2016jointly}, and generation of a trajectory for the robot's end-effector based on robot's dynamics model~\cite{ritz2012cooperative,lampariello2011trajectory, kocc2018online} or model-free RL~\cite{tebbe2021sample, yang2021ball} to catch the ball at the predicted interception point.
An alternative approach~\cite{gao2020robotic} is to learn an end-to-end policy in simulation that directly takes the camera's RGB image as input, followed by fine-tuning in the real world~\cite{abeyruwan2022sim2real,buchler2022learning}. However, for quadrupeds, the previous model-based methods which require accurate modeling of the ball and the robot will be hard to utilize due to the complexity of the dynamics models, while previous model-free RL methods have not been applied to control such dynamic legged robot for manipulation tasks.

\subsubsection{Dynamic Locomotion Control for Quadrupeds}

In recent years, there have been considerable advances in legged robot hardware and control algorithms that enable quadrupedal robots to preform highly dynamic locomotion maneuvers, such as jumping~\cite{peng2020learning,nguyen2019optimized,gilroy2021autonomous,park2021jumping,nguyen2021contact,bogdanovic2022model} or running~\cite{kim2019highly,ji2022concurrent,margolisyang2022rapid}, in the real world. 
One approach is to utilize an optimal control framework with the robot's dynamics models, which can be the robot's full-order models and optimized offline~\cite{nguyen2019optimized,gilroy2021autonomous,nguyen2021contact}, or simplified models and deployed online~\cite{kim2019highly,park2021jumping}. 
Another approach is to leverage model-free deep RL to train the quadrupedal robots through trail-and-error in simulation first and then transfer to the real robot~\cite{peng2020learning,bogdanovic2022model,ji2022concurrent,margolisyang2022rapid}. 
However, most previous work only focuses on a specific dynamic locomotion skill without attaining a more diverse repertoire of maneuvers based on learned skills to achieve a longer horizon task, such as jumping while tracking different swing leg trajectories to intercept a ball.

\subsubsection{Legged Robot Soccer}

Developing robots that can one day compete with humans in soccer games has been an enduring goal in the robotics community, and a notable soccer robot game is RoboCup~\cite{veloso2012video}.
Related to the goalkeeping problem of this work, there are some efforts to develop an intelligent goalkeeper using holonomic wheeled robots~\cite{lausen2003model, cunha2011ball, cooksey2016opponent}.
However, most previous work only consider the robot moving in 2D plane to intercept a ball rolling on the ground at low speeds~\cite{cunha2011ball,lausen2003model}. 
Intercepting balls in a 3D and at high speeds, like a flying ball with a speed up to 8 m/s, as in this work, has not been studied in robot soccer. 
Legged robots, such as humanoid robots and quadrupedal robots, are also used in RoboCup, but most presented soccer skills by legged robots, such as shooting~\cite{cherubini2010policy}, kicking~\cite{teixeira2020humanoid}, and goalkeeping~\cite{masterjohn2015regression}, are based on rule-based motion primitives due to their challenging dynamics.
Most recently, by leveraging deep RL, a quadrupedal robot demonstrates the capacity to dribble a soccer ball to a target at a low walking speed~\cite{bohez2022imitate}, and a quadruped is also trained to precisely shoot a soccer ball to a random given target while the robot is standing with a single shooting skill~\cite{ji2022hierarchical}.  
However, enabling legged robots to play soccer while performing multiple highly dynamic locomotion skills, such as using jump and dive skills, and precise ball manipulation has not yet been demonstrated.

\subsection{Contributions}
The core contribution of this work is the creation of an agile and dynamic quadrupedal goalkeeper for robot soccer. 
This work presents one of the first solutions that combines both highly dynamic locomotion and precise object interception (manipulation) on real quadrupedal robots by using a hierarchical reinforcement learning framework. 
The proposed method allows quadrupeds to track parametric trajectories with its end-effector(s) while engaging in dynamic locomotion maneuvers. 
The hierarchical framework is used to learn and compose a diverse set of low-level locomotion skills, and to select the most appropriate skill and motion for the robot to intercept a flying ball. 
We show that our system can be used to directly transfer dynamic maneuvers and goalkeeping skills learned in simulation to a real quadrupedal robot, with an 87.5\% successful interception rate of random shots in the real world.
We note that human soccer goalkeepers average around a 69\% save rate, \cite{saverate}.  Although, this is against professional players shooting towards regulation sized goals, we hope this paper takes us one step closer to enabling robotic soccer players to compete with humans in the near future.

\label{sec:Introduction}

\section{Hierarchical RL Framework for Goalkeeping Task with Multi-Skills}
\label{sec:framework}

\begin{figure*}
     \centering
     \includegraphics[width=0.7\textwidth]{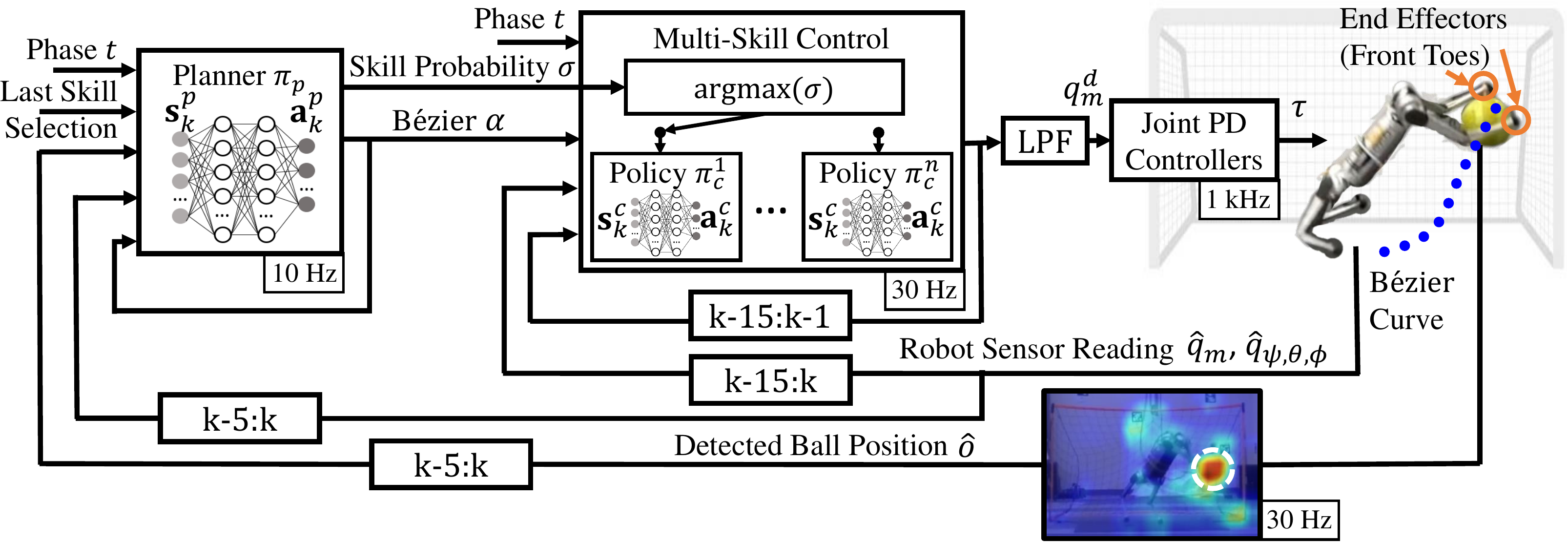}        \caption{Proposed hierarchical reinforcement learning framework for creating a quadrupedal robotic goalkeeper. We firstly develop a set of locomotion control policies for different skills, such as sidestep, dive, and jump. The locomotion control policies are designed to follow random parametric B\'ezier curve using the robot end-effectors (swing front toes). The controller outputs desired motor position at $30$ Hz for the joint-level PD controller to generate motor torques, after passing through a Low Pass Filter~(LPF)~\cite{peng2020learning,li2021reinforcement}. A motion planner running at $10$ Hz is developed on top of the multiple skill-specific controllers to select the specific skill to perform as well as the desired end-effector trajectory for the controller to track. The goal of the planner is to enable the robot to intercept the ball via its body before the goal. The controllers and planners are trained by RL and the ball position is detected by a deep neural network using a RGB-Depth camera ($30$~Hz).}
    \label{fig:framework}
    \vspace{-0.4cm}
\end{figure*}

\begin{figure}
     \centering
     \includegraphics[width=0.8\linewidth]{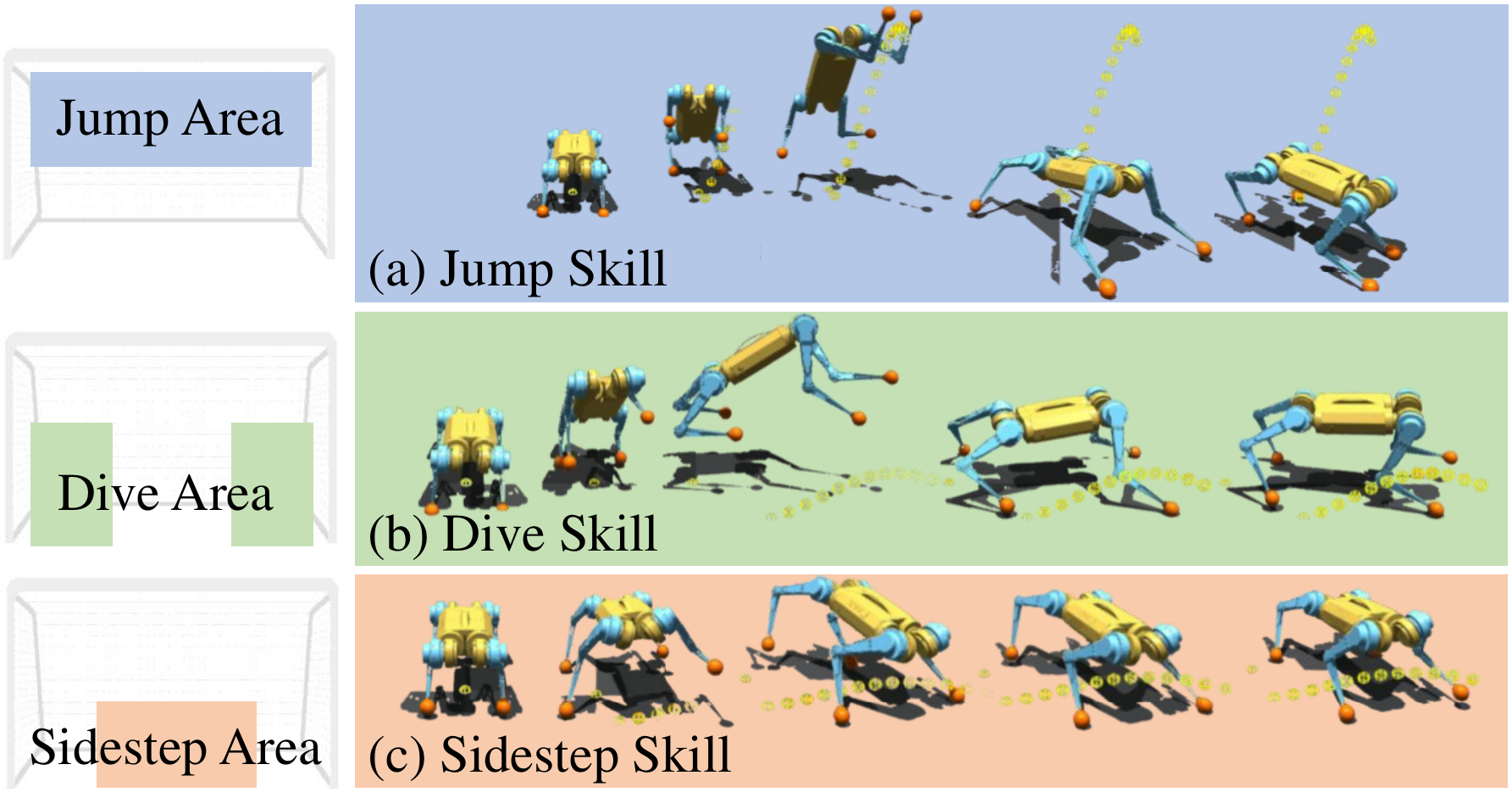}
     \caption{Three different locomotion skills for goalkeeping. The robot can use different skills to cover different regions of the goal.}
    \label{fig:multiskill}
    \vspace{-0.4cm}
\end{figure}

In this section, we introduce the Mini Cheetah robot which is the experimental platform for this work. 
We also provide a brief overview of the framework for developing goalkeeping skills as illustrated in Fig.~\ref{fig:framework}.

\subsection{The Mini Cheetah Quadrupedal Robot}
As shown in Fig.~\ref{fig:intro}, Mini Cheetah~\cite{kim2019highly} is a quadrupedal robot having a weight of $9$~kg and height of $0.4$~m when it is fully standing. It has $12$ actuated motors $q_m\in \mathbb{R}^{12}$ and a $6$ degree-of-freedoms (DoFs) floating base, representing its translational $q_{x,y,z}$ (sagittal, lateral, and vertical) positions and orientation $q_{\psi,\theta,\phi}$ (roll, pitch, yaw), respectively.

\subsection{Locomotion Skills for Goalkeeping}~\label{subsec:skills}
Inspired by human goalkeepers, we propose a collection of skills for intercepting a ball flying to different regions of the goal, as illustrated in Fig.~\ref{fig:multiskill}.
The main concern underlying the design of goalkeeping locomotion skills is that the robot needs to react very quickly, since the total timespan of a ball's ballistic trajectory is typically under $1$ sec.
Therefore, from an initial standing pose in the middle of the goal, the robot needs to perform very dynamic maneuvers to intercept the ball.
To accomplish this, our system uses three locomotion skills: \textit{sidestep}, \textit{dive}, and \textit{jump} to cover different goal regions.
\subsubsection{Sidestep}
During a sidestep, the robot takes a quick step in the lateral direction to intercept the ball when it is rolling on the ground or flying toward the goal at a low attitude.
Depending on the size of the step, the robot may only need to swing up one of its front leg while the rest can remain in the stance phase.
But for larger steps, the stance legs may also need to leave the ground, resulting in a small sideways hop. 
However, the sidestep skill may not be able to cover regions that are farther away from the robot, such as the lower corners of the goal or the upper regions.
\subsubsection{Dive}
The dive skill is based on quadrupedal jumping behaviors~\cite{gilroy2021autonomous}, which allows the robot to cover a larger area of the goal. 
Using the dive skill, the robot should first pitch its body up onto the rear legs, then turn to the lateral side towards the direction that the ball is traveling, extend its two swing legs to reach the ball, and finally land back on its feet.
This skills enables the robot to quickly block the lower corners of the goal. 
During the dive, the rear legs may or may not leave the ground, depending on how far the robot needs to travel.
\subsubsection{Jump}
Similar to the dive skill, the jump skill also requires the robot to pitch up its body and swing its front legs upwards as fast as possible. 
But for the jump, the robot also needs to extend its swing legs higher to intercept the ball, when it is traveling towards the upper regions of the goal.
To perform this dynamic jump, the robot needs to use its rear legs to push off the ground in order to extend its front legs to reach to higher regions. 
After the ball has been intercepted, the robot needs to reconfigure itself in the air to a more stable landing pose.

For each of these three skills, a nominal reference motion for the Mini Cheetah is manually authored using a $3D$ animation tool~\cite{li2020animated}. Each reference motion starts in the same default standing pose. RL is then used to train controllers to perform each skill by imitating the respective reference motion.

\subsection{Parameterize Multiple Skills using B\'ezier Curves}~\label{subsec:bezier}
Our robot leverages the above-mentioned three locomotion skills to reach different commanded locations in front of the goal in order to intercept the ball.
Inspired by previous work~\cite{li2021reinforcement,ji2022hierarchical}, we use B\'ezier curves to parameterize the desired robot motion.
A B\'ezier curve parameterized by B\'ezier coefficients $\alpha$ and is defined as a function of phase $t$, $B_\alpha (t)$~\cite[Sec.II-C]{ji2022hierarchical} where $t \in [0,1]$ is the normalized time w.r.t. a total duration $T$ of the entire trajectory. 
Similar to~\cite{ji2022hierarchical}, the B\'ezier curve represents the desired trajectory for the robot's end-effectors, which is designated to be the toes.
In the \textit{sidestep} skill, the end-effector is designated to be the toe of the robot's swing leg, either the front right or left toe.  
However, for the skills that need to use two swing legs to catch the ball, like \textit{jump} and \textit{dive}, the B\'ezier curve specifies a trajectory for the center of the two end-effectors (two front swing toes).
In this work, we choose to use $5$ control points in $3D$ space, therefore, $\alpha \in \mathbb{R}^{5\times3}$. 
The duration $T$ of each skill is set to be $0.5$ seconds.
In this way, we can parameterize and represent different robot's end effector trajectory to reach different locations by using different B\'ezier coefficients.

\subsection{Hierarchical Reinforcement Learning Framework}
We now present the proposed hierarchical RL framework for controlling a quadrupedal robotic goalkeeper shown in Fig.~\ref{fig:framework}.

For each goalkeeping locomotion skill, we choose to train a control policy to have joint-level commands for the robot using model-free RL.
This enables the robot to mimic the nominal reference motion while tracking a large range of randomized end-effector trajectories represented by parametric B\'ezier curves. 
This process produces multiple control policies ($\pi^1_c \dots \pi^n_c$) for different goalkeeping skills. 
In this work, we trained three control policies for the \textit{sidestep}, \textit{dive}, and \textit{jump} skills, and each controller runs at $30$ Hz.
Since each skill performs a distinct behavior, training individual skill-specific control policies avoids the challenge of training a single multi-task policy for all skills. 
To select the appropriate skill for different scenarios representing the ball traveling towards different regions of the goal, we train a high-level planning policy $\pi_p$ using model-free RL to select a desired skill depending on the detected ball position and the robot's current states. 
The planning policy operates at $10$~Hz. 
It also specifies the desired B\'ezier curve for the chosen controller to track in order to intercept the ball.
The ball position is obtained by an external camera and a deep supervised learning algorithm YOLO~\cite{wang2022yolov7}, without building a ball state estimator. 

We first train each control policy in simulation, and test them extensively on the real robot with zero-shot transfer. 
After obtaining reliable controllers in the real world, we then use the same control policies to train the planning policy in simulation and then directly deploy the entire pipeline in the real world. 
This approach decouples the complex goalkeeping task into the locomotion control and manipulation planning problems, and solves each problem separately.

\section{Dynamic Locomotion Control using Deep RL}
In this section, we detail our framework for training control policies for each goalkeeping skill.

\subsection{Training Environment}
The control policies are trained using a simulation of the Mini Cheetah in Isaac Gym~\cite{makoviychuk2021isaac}, a GPU-accelerated dynamics simulator. For a skill-specific controller $c$, at each timestep $k$ of an episode, the robot takes an
action $\mathbf{a}^c_k$ based on its current observations $\mathbf{s}^c_k$, and the environment transits to the next state and provides a reward $r_{c,k}$. The RL is used to maximize the expected accumulated reward over the course of an episode.   

\subsubsection{Action Space}
The skill-specific control policy outputs a $12$-dimensional action that specifies target motor positions for each joint $q^d_m$ and is used by joint-level PD controllers to compute motor torques $\tau$. 

\subsubsection{Observation Space}
As shown in Fig.~\ref{fig:framework}, the policy's observations consist of three components. 
The first component is the desired end-effector trajectory for the robot to track,  represented by a set of B\'ezier coefficients and normalized phase $t$ w.r.t. the $0.5$-sec-timespan of the motion.
The second component is the robot's raw sensor reading, which consists of the measured motor positions $\hat{q}_m$ and base rotation $\hat{q}_{\psi,\theta,\phi}$. 
We also include a history of the sensor readings in the past $15$ timesteps, which corresponds to a time window of around $0.5$ seconds. 
The last component is the robot's actions ($q^d_m$) in the past $15$ timesteps.
Incorporating a history of the robot's actions and sensor readings provides the policy with some information necessary for inferring the dynamics of the system, which can be vital for sim-to-real transfer~\cite{li2021reinforcement}.

\subsection{Reward Formulation}
The reward $r_{c,k}$ at timestep $k$ is designed to encourage the control policy to track the given B\'ezier curve for the robot's end-effector, while smoothly following the skill-specific reference motion and maintaining gait stability. 
The reward function contains three main components: 
\begin{equation}
    r_{c, k} = 0.5r_{c,k}^E + 0.3r_{c,k}^I + 0.2r_{c,k}^S,
\end{equation}
where $r_{c,k}^E$ represents the robot's end-effector tracking term, $r_{c,k}^I$ is the imitation term, and $r_{c,k}^S$ is a smoothing term. 
The end-effector tracking term has the highest weight while the smoothing term has the lowest, which is to put higher priority on tracking the desired B\'ezier curve. 
Next, we define an abstract reward function 
\begin{equation}\label{eq:reward}
    r(u,v) = \exp[ \rho ||u-v||^2_2] ,
\end{equation}
which calculates the normalized distance between vector $u$ and $v$, with $\rho > 0$ being a hyperparameter. 
With this, the $r^E_{c,k}$ term can be decomposed into:
\begin{equation}
    r_{c,k}^E = 0.8r(B_\alpha(t), x_{e, k}) + 0.2r(\dot{B}_\alpha(t), \dot{x}_{e, k}),
\end{equation}
where $B_{\alpha}(t)$ is the desired end-effector position evaluated at phase $t$ at current timestep $k$ and $\dot{B}_{\alpha}(t)$ is the derivative of the curve at phase $t$ which represents the desired end-effector velocity. 
The $x_{e,k}$ is robot's end-effector position in Cartesian space while $\dot{x}_{e,k}$ is its velocity. 
We empirically found that adding the end-effector velocity tracking term can improve the smoothness of the end-effector trajectory. 

Similarly, the imitation reward $r_{c,t}^I$ encourages the robot to mimic the skill-specific reference motion, and consists of $3$ terms: the motor tracking reward $r(q_m, q^r_m)$, robot base height following reward $r(q_z, q^r_z)$, and base orientation tracing reward $r(q_{\psi, \theta, \phi}, q^r_{\psi, \theta, \phi})$, by calculating the distance between robot's current values and the one from reference motion ($q^r_{m, z, \psi, \theta, \phi}$) at each timestep. 
Similarly, $r_{c,k}^S$ is the smoothing reward and consists of $r(\dot{q}_{\psi, \theta, \phi},0)$, $r(\ddot{q}_{\psi, \theta, \phi},0)$, and $r(\tau,0)$ to minimize the robot base rotational velocity, rotational acceleration, and torque consumption, respectively.  
By incorporating the smoothing reward, we can effectively minimize nonessential angular movement that hampers robot stability, and energy consumption throughout the motion.

\subsection{Motion and Dynamics Randomization}
With the skill-specific nominal reference motion unchanged, the desired end-effector trajectory is randomized by adding a random change of B\'ezier coefficients $\tilde{\alpha}$ to a nominal set of B\'ezier coefficients $\bar{\alpha}$, \textit{i.e.}, $\alpha=\bar{\alpha}+\tilde{\alpha}$. The coefficients change $\tilde{\alpha}$ of each control point is uniformly sampled from a large range, especially in the lateral direction to cover farther sides of the goal. 
This then allows each policy to track a large variety of end-effector trajectories defined by the randomized $\alpha$.

During training in simulation, dynamics parameters are randomized to facilitate transfer from simulation to the real world. Similar to previous work~\cite{ji2022hierarchical}, we randomize the robot's link mass, inertia, and center of mass in a given range to mitigate modeling error. We also randomize the PD gains of the motors to accommodate modeling errors in the motor dynamics. We also simulate sensor noise and delay using a similar randomization range as ~\cite{ji2022hierarchical}. 
Ground friction plays a critical role during sim-to-real transfer for the jump and dive skills where the robot needs to use its rear legs to push off the ground. 
Therefore, the friction coefficient is randomized within a large range of $[0.5, 4.5]$. 
The ground restitution is also randomized from $0$ to $0.5$ for better adaptation to soft padded ground. 
Additionally, a random $6$ DoFs wrench (force varying from $-1$ to $1$ N and torque is from $-3$ to $3$ Nm) is applied on the robot base to perturb the robot and improve the robustness of the policy.
Specifically, we found that a large perturbation in roll direction, ranging from $-12.5$ Nm to $12.5$ Nm, helps to prevent the robot from rolling over when jumping sideways.

\subsection{Episode Design and Training}
Each episode lasts for $300$ timesteps, corresponding to $10$ seconds and consists of three stages. 
In the first stage, the robot starts in a nominal standing pose. After a random span of time has elapsed, a set of desired B\'ezier coefficients is randomized and given to the policy.
The robot is then supposed to follow the desired end-effector trajectory, while also imitating the reference motion of the specific skill.
After the desired goalkeeping maneuver has been completed, the reference motion switches to a nominal standing motion to encourage the robot to recover back to the nominal pose. 

Two early termination conditions are applied during training. First, the episode is terminated whenever there is an unsafe behavior, like the robot falling over or one leg colliding with another part of the body, resulting in zero reward for all future timesteps. Secondly, a large deviation from the commanded curve for the end-effectors also results in termination of the episode. 
This criteria further encourages the robot's end-effector to stay as close as possible to the desired trajectory. 
We found that early termination leads to faster training and more precise tracking of different end-effector trajectories.  

All policies are trained with Proximal Policy Optimization (PPO)~\cite{schulman2017proximal}. 
The actor and critic networks are modeled by separate MLPs with hidden layers consisting of 512, 256, 128 units, with ELU activation functions.

\section{Multi-Skill Motion Planning using Deep RL}
\label{sec:planning}
Each obtained skill-specific locomotion control policy is first tested thoroughly on the robot hardware. 
After the sim-to-real test, these control policies can then be used to train a planning policy that enables the robot to intercept the ball by performing dynamic locomotion maneuvers after examining the detected ball position and the current robot states. 

\subsection{Training Environment} 
The planning policy is also trained in Isaac Gym with the Mini Cheetah driven by its controller and a rigid ball. The goal in the simulation is sized $1.5$ m wide and $0.9$ m high, and the robot is placed $0.2$ m in front of the goal line. 
\subsubsection{Action Space}
As shown in Fig.~\ref{fig:framework}, the planning policy outputs the desired skill type to perform, and the desired end-effector trajectory for the selected controller to track. 
Therefore, the action of the planner $\mathbf{a}_p$ contains two parts: the skill type and desired B\'ezier coefficients $\alpha^d$.
For $n$ skills, the planner outputs skill selection probabilities $\sigma \in \mathbb{R}^{n+1}$, and the desired skill type can be determined by finding the argmax of $\sigma$. The extra skill utilizes a time-invariant standing pose with a PD controller. In this work, $n=3$. 
Although there are $5$ control points to construct the B\'ezier curve we used in the control policy, the planner outputs only $4$ of them, excluding the first control point, \textit{i.e.}, $\alpha^d \in \mathbb{R}^{4\times3}$. 
The first control point is always overwritten by a nominal initial position of the robot's end-effector to reduce the learning complexity.

\subsubsection{Observation Space}
The observation of the planning policy $\mathbf{s}^p_k$ consists of four components.
The current (at timestep $k$) detected ball position in global frame $\hat{o}_k$. Instead of developing ball position estimation and prediction system, a history of the ball positions in the last $6$ timesteps is given to the planning policy.
This is helpful for the planning policy to filter out the noisy readings from the camera in the real world, to estimate its velocity and high order information, and to infer the ball's future trajectory, implicitly.
The second component is the current robot sensor reading, including motor angles $\hat{q}_m$ and base orientation $\hat{q}_{\psi,\theta,\phi}$, to enable the policy to be aware of robot's current states. 
Similar to the control policy, a history ($6$ timesteps) of the robot feedback and actions is provided to the planning policy to implicitly learn to estimate dynamics. Please note that, although the planner runs at $10$~Hz, all the feedback history is sampled at $30$ Hz which is the higher frequency that the controller uses.
Moreover, the last planner action, which is the previous skill selection and desired B\'ezier coefficients are also input to the policy.
By informing the planning policy the previous skill and motion selection, together with the robot current states, planner can learn to avoid making sudden but infeasible changes of the skills, such as commanding the robot to perform sidestep skill while the robot is in the air with a previous selected jump skill. 
Finally, the phase $t$ of the performed skill is also included.

\subsection{Reward Formulation}
The reward of the planning policy is designed to perform a "save" by intercepting the flying ball with the robot's body, robot's end-effectors or trunk. To facilitate this, the reward is formulated as: 
\begin{equation}
r_{p,k} = b_k (r_{p,k}^{\dot{o}} + 0.6 r_{p,k}^{x_{e}} + 0.2 r_{p,k}^{x_e^d} + 0.2 r_{p,k}^{\alpha^d}), \end{equation} %
where 
\begin{equation}
b_k = \begin{cases}
1,~~~\text{if}~~||o_{k} - x_{e,k}|| \leq 0.3 \\
0,~~~\text{otherwise}
\end{cases}  
\end{equation}
is a binary variable indicating if the current ball position $o_k$ is close enough to robot's current end-effector position $x_{e,k}$. 
In this way, we only calculate the reward when the ball is close to the robot.  
Since the robot is only allowed to perform one save in each episode, by such a sparse reward design, the robot is encouraged to just stand still without performing other goalkeeping skills until the ball is nearby for a better return over the entire episode.  

The dominating reward term $r_{p,k}^{\dot{o}}$ represents the reward for the ball velocity.
It is set to $1$ if the ball speed $||\dot{o}||_2$ is zero and to $0$ otherwise. By this term, we encourage the robot to stop the ball.
Furthermore, the reward $r_{p,k}^{x_e}$ stimulates the robot to minimize the distance between the robot's end-effector position $x_{e}$ to the ball by $r(x_{e,k}, o_k)$ where $r(\cdot, \cdot)$ is defined via~\eqref{eq:reward}.
Similarly, $r_{p,k}^{x_e^d}$ incentivizes the planning policy to enable the desired end-effector trajectory to track the ball by $r(B_{\alpha^d}(t),o_k)$.
Please note that, since the dominating term is to stop the ball ($r_{p,k}^{\dot{o}}$), the robot can also receive reasonably good reward if it uses other parts of the body other than the end-effectors, such as the trunk, to save the ball.
Finally, the smoothing term $r_{p,k}^{\alpha^d} = r(\alpha^d,0)$ is introduced to encourage the planner to regularize the desired B\'ezier coefficients $\alpha^d$ to prevent having fluctuating curves.

Please note that we did not directly include a reward to minimize the energy consumption of the robot. The reason is that the motor torques have a much larger variance between skills than within skills, and thus, the robot may easily learn to keep using a skill with low energy consumption, such as sidestep, but overlook the ball saving task.

\subsection{Early Termination Condition}
Besides the reward design, the termination conditions to end the episode earlier are critical to enable the robot to save the ball in front of the soccer goal. 
We terminate the episode to prevent the agent from having a future return if the ball flies into the goal area. 
This can stimulate the robot to try its best to save the ball instead of adopting conservative behaviors like just standing where the robot can still receive some rewards such as the smoothing reward. 
Further, the episode will be terminated if the robot falls over. 
This can prevent the planning policy from outputting infeasible end-effector trajectories or commanding wrong skills, like selecting a sidestep skill while the robot is in the air, because these will cause the locomotion controllers to fail.

\subsection{Episode Design, Domain Randomization, and Training}~\label{subsec:planning_episode}
Each episode lasts for $90$ timesteps ($3$ seconds) in total. 
Upon reset, the ball's initial $3D$ position and velocity in the transverse plane is randomly sampled. 
The target position for the ball is sampled within the goal area, and the initial vertical speed is obtained accordingly.

We again leverage domain randomization to improve the robustness of the policy. 
In addition to those during training the control policy, we also applied Gaussian noise with zero mean and $0.05$ m standard deviation and constant delay randomized from $80$ to $100$ ms for the perceived ball positions.
The constant delay is key to succeed in the sim-to-real transfer because there is substantial delay mainly from the camera.

The planning policy has separated and identical actor and critic networks that consist of a 2-layer MLP with 256 and 128 hidden dimensions with ELU as the activation function. 
The last layer of the actor network is followed by two action heads, one for continuous Bezier coefficients, and one for categorical skill selection action. 
A $tanh$ and a $softmax$ activation are appended to the continuous and discrete heads respectively. 
The planning policy is also optimized by PPO. 
We found that the planning policy trained in simulation can be directly deployed in the real world without finetuning with a soft ball unlike what was done in~\cite{ji2022hierarchical}.

\section{Results}
After having developed all components of the proposed framework (Fig.~\ref{fig:framework}), we now validate the proposed framework in simulation and experiments. Experiments are also recorded in the accompanying video (\url{https://youtu.be/iX6OgG67-ZQ}).

\subsection{Simulation Validation}

\begin{figure}
     \centering
     \begin{subfigure}[b]{0.15\textwidth}
         \centering
         \includegraphics[width=\textwidth]{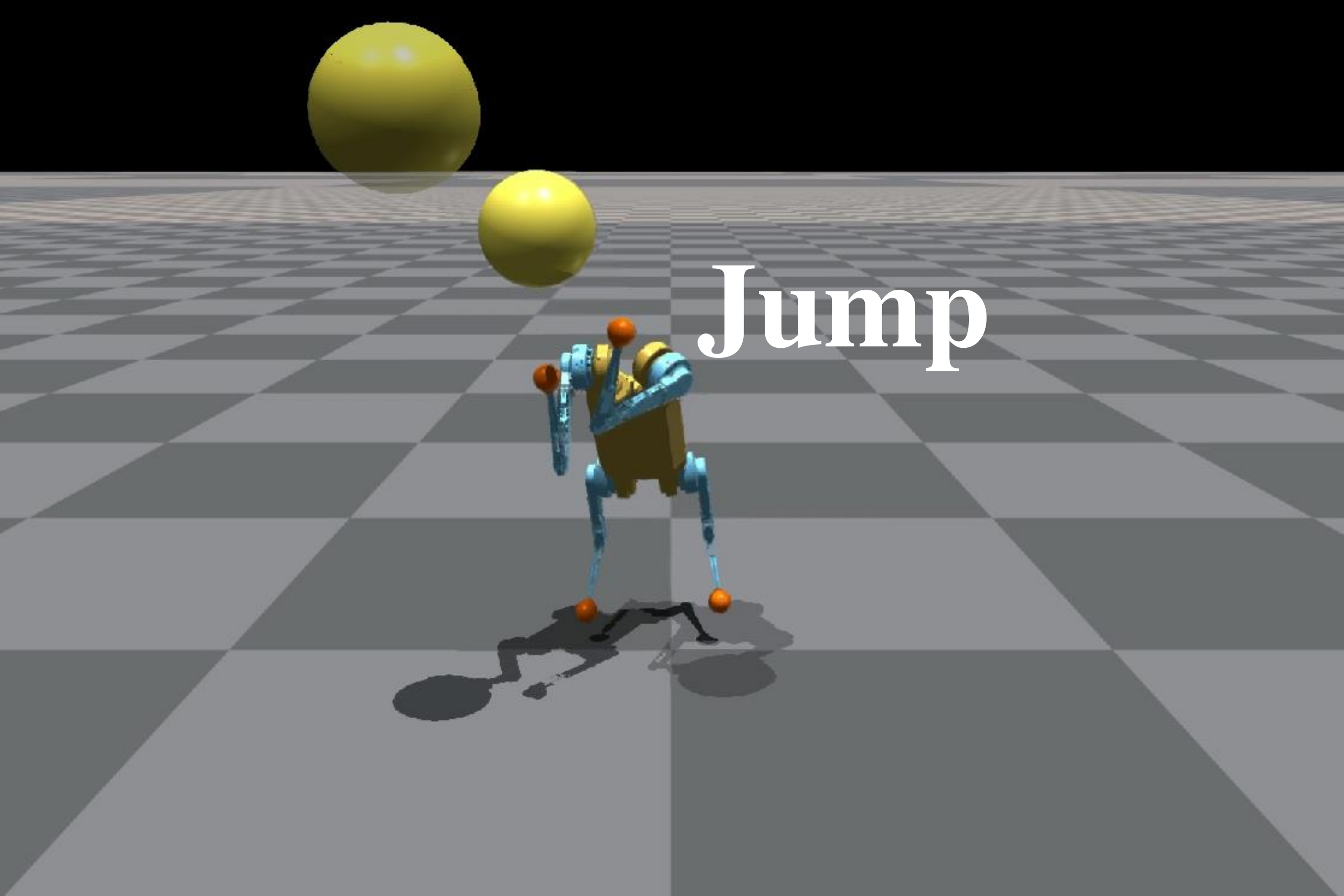}
     \end{subfigure}
     \hfill
     \begin{subfigure}[b]{0.15\textwidth}
         \centering
         \includegraphics[width=\textwidth]{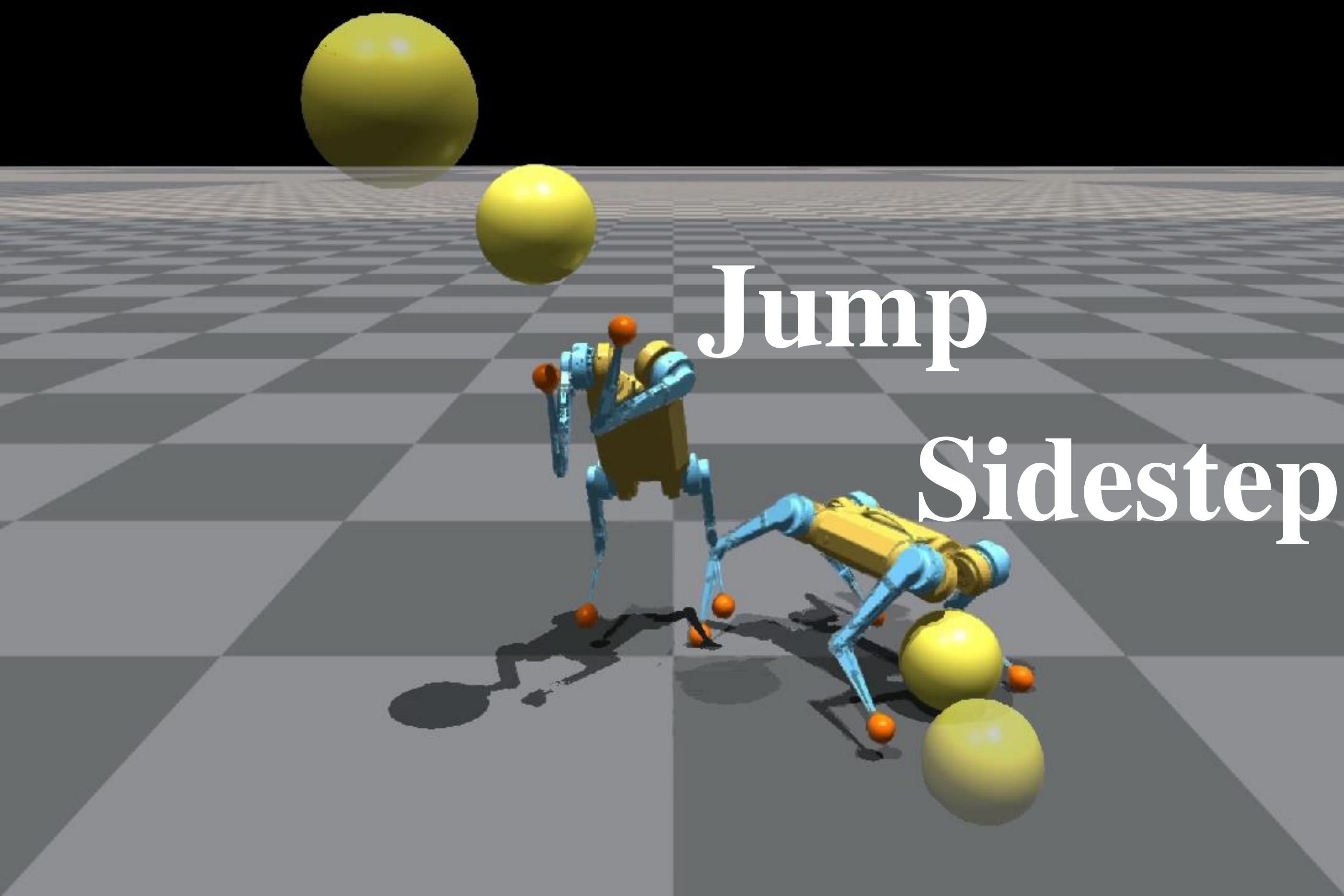}
     \end{subfigure}
     \hfill
     \begin{subfigure}[b]{0.15\textwidth}
         \centering
         \includegraphics[width=\textwidth]{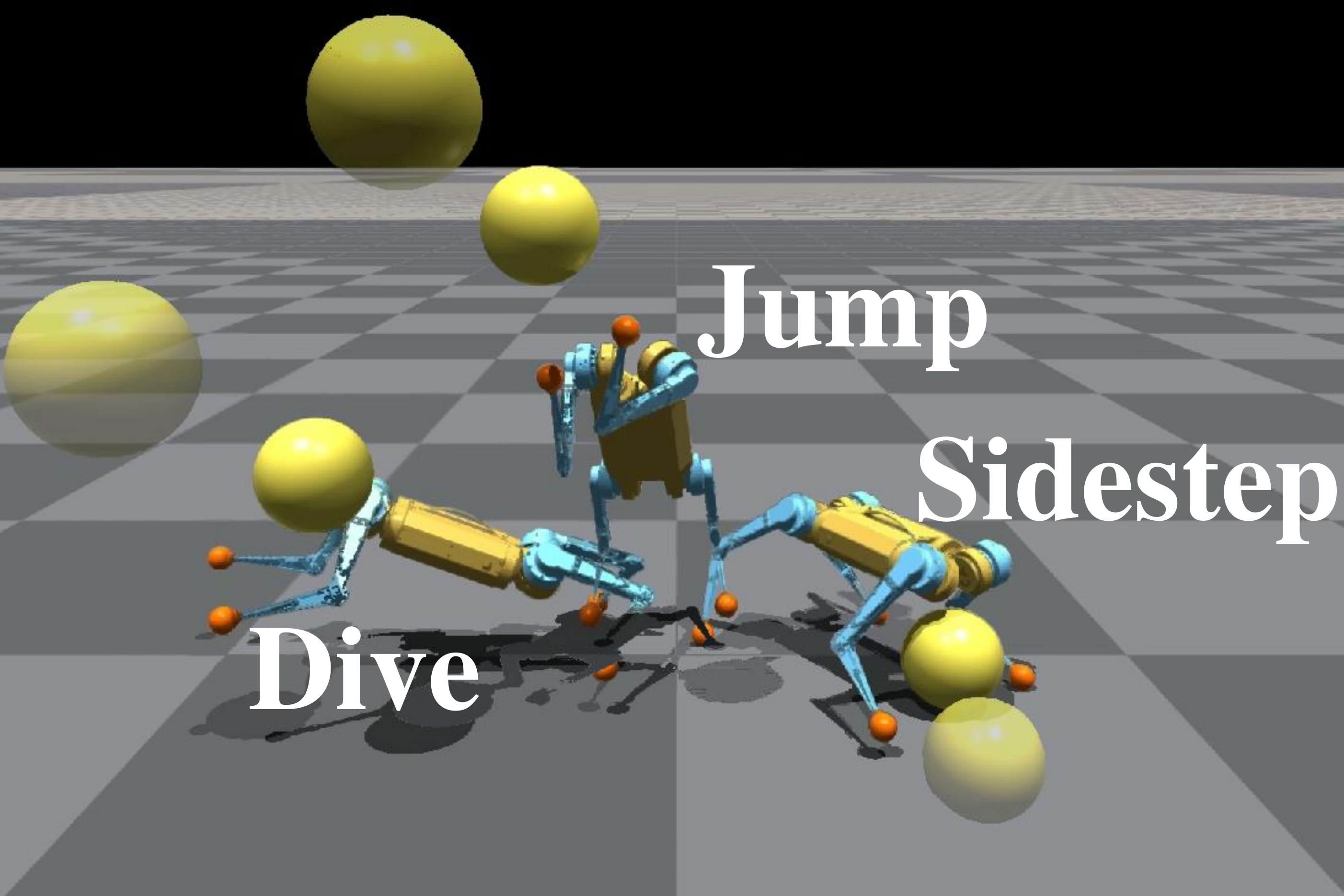}
     \end{subfigure}
     \newline
     \begin{subfigure}[b]{0.48\textwidth}
         \centering
         \includegraphics[width=\textwidth]{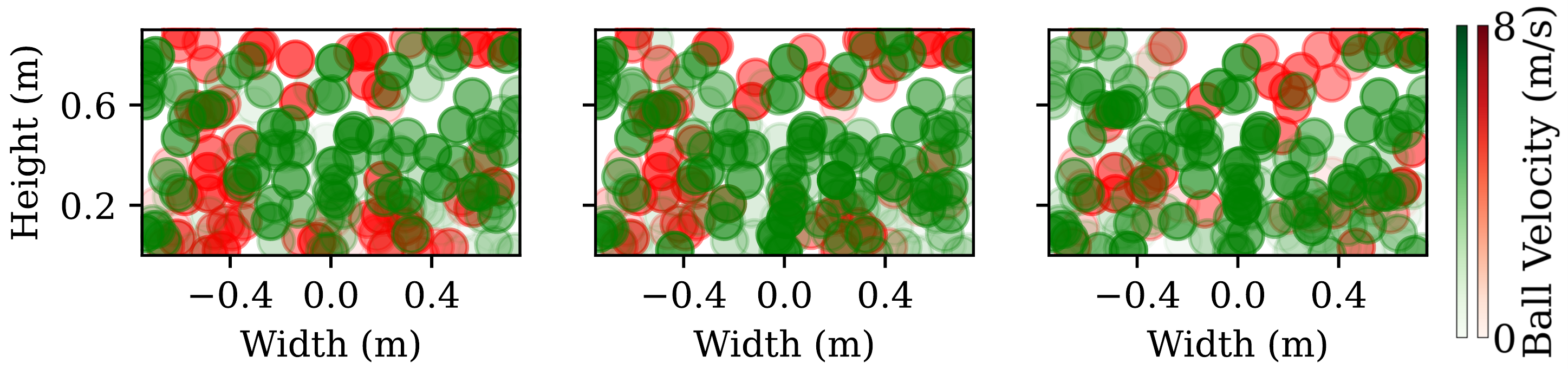}
     \end{subfigure}
     \begin{subfigure}[b]{0.15\textwidth}
         \centering
         \caption{1-Skill Planner}
         \label{fig:1skillsim}
     \end{subfigure}
     \hfill
     \begin{subfigure}[b]{0.15\textwidth}
         \centering
         \caption{2-Skill Planner}
         \label{fig:2skillsim}
     \end{subfigure}
     \hfill
     \begin{subfigure}[b]{0.15\textwidth}
         \centering
         \caption{3-Skill Planner}
         \label{fig:3skillsim}
         \end{subfigure}
     \captionsetup{singlelinecheck = false, justification=justified}
        \caption{Snapshots and shot interception map in simulation with more skills added from left to right. The map represents the goal region. Green records a goal save while red is a goal (miss). Darker colors indicates faster ball speeds. The snapshots visualize how the planner leverages the new skills, and the shot interception map quantitatively illustrate the benefits of adding each skill. Note that the failing corner cases are noticeably reduced by adding the second sidestep skill in \ref{fig:2skillsim}, and further reduced by the third dive skill in \ref{fig:3skillsim}. The goal saving rates are 65.09\%, 72.46\%, and 78.11\%, respectively.}
        \label{fig:accumapsim}
    \vspace{-0.4cm}
\end{figure}

We firstly evaluate the performance of the proposed multi-skill framework in simulation. 
Specifically, we compare the ball saving rate using different number of locomotion skills. They are: \textit{jump-only}, \textit{jump-sidestep}, \textit{jump-dive-sidestep} (ours). 
We trained three planners using the same method introduced in Sec.~\ref{sec:planning} for these different combination of skills, respectively. We denote these planners as \textit{1-skill}, \textit{2-skill}, and \textit{3-skill} (ours) planner accordingly. 
Each method is repeated 200 times with randomized scenarios as specified in Sec.~\ref{subsec:planning_episode}. 

As shown in Fig.~\ref{fig:accumapsim}, comparing with the planners with more than one skill, the \textit{1-skill} planner achieves a comparable saving rate with flying balls, but misses almost all of the ground rolling balls, resulting in $65.09\%$ saving rate. 
Such a result demonstrates that using a single skill is not sufficient for the goalkeeping task. 
While the \textit{2-skill} planner can catch most of the balls ($72.46\%$), however, there are two notable corner cases on the lower left and right that hampers the performance. 
We discovered that in these cases a majority of the balls travel underneath the robot when the jumping skill is activated, and can not be reached by the robot when it steps to the side. 
This problem highlights the necessity of dive skill to save the ball flying to these regions. Considering the dive skill, the \textit{3-skill} planner shows the best saving rate of $78.11\%$.

\subsection{Experiments}
We now deploy the proposed framework on the Mini Cheetah robot to save soccer goals in the real world.

\subsubsection{Experiment Setup}
We set up a mini penalty field to conduct the experiments, as shown in Fig \ref{fig:intro}, with a $1.5m \times 0.9m$ goal. 
The robot is placed at the center with its rear feet $0.1$~m in front of the goal line. The soccer ball, which is size 3, is either kicked or thrown roughly $4$~m in front of the robot with a random initial speed towards a random target in the goal. We set an external RGB-Depth camera (Intel RealSense D435i) placed $6$~m away from the goal line. 
We also setup a Motion Caption System~(MoCap) with markers on the robot's trunk, front toes, and soccer ball, to evaluate the tracking performance of the locomotion controllers. Please note that our system does not require accurate measurement from the external MoCap.

\subsubsection{Performance of Skill-specific Locomotion Controllers}
\begin{figure}
     \centering
     \begin{subfigure}[b]{0.45\textwidth}
         \centering
         \includegraphics[width=\textwidth]{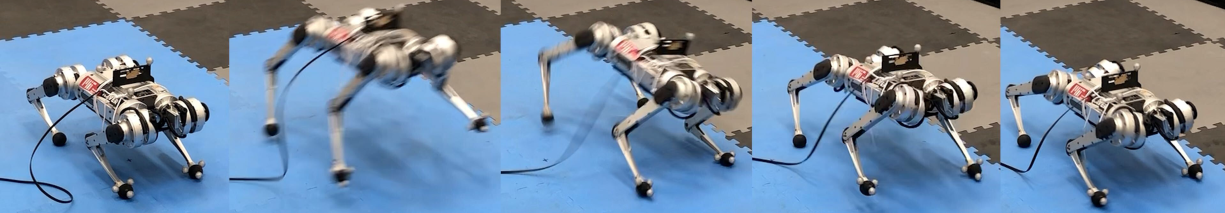}
         \caption{Sidestep}
         \label{fig:swmocap}
     \end{subfigure}
     \begin{subfigure}[b]{0.45\textwidth}
         \centering
         \includegraphics[width=\textwidth]{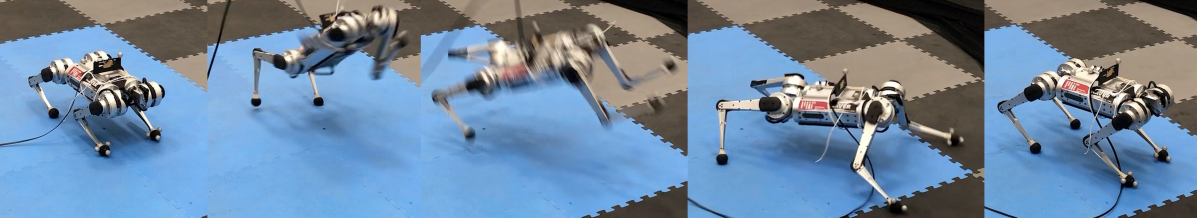}
         \caption{Dive}
         \label{fig:divemocap}
     \end{subfigure}
    \begin{subfigure}[b]{0.45\textwidth}
         \centering
         \includegraphics[width=\textwidth]{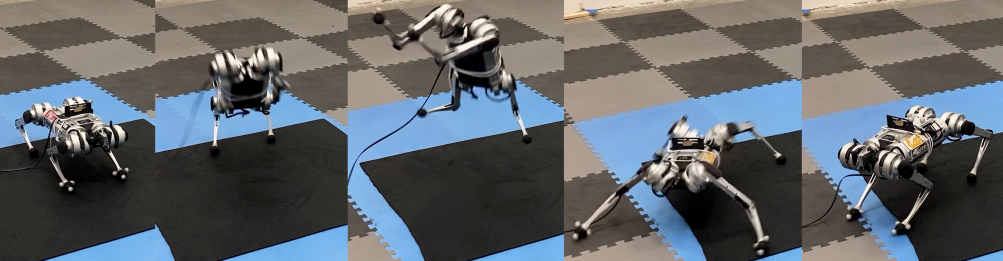}
         \caption{Jump}
         \label{fig:jumpmocap}
     \end{subfigure}
    
        \captionsetup{singlelinecheck = false, justification=justified}
        \caption{Experiments with control policies for different skills. The policy is able to directly transfer to the hardware. As designed in Sec.~\ref{subsec:skills}, we can observe that the dive skill \ref{fig:divemocap} is able to reach a significantly larger range horizontally than sidestep \ref{fig:swmocap}, while the jump skill \ref{fig:jumpmocap} can produce a notable period of flight time, swing the front legs to cover more upper-altitude area, and land safely.}
        \label{fig:controltrack}
        \vspace{-0.4cm}
\end{figure}

\begin{figure}
     \centering
    \begin{subfigure}[b]{0.48\textwidth}
         \centering
         \includegraphics[width=\textwidth]{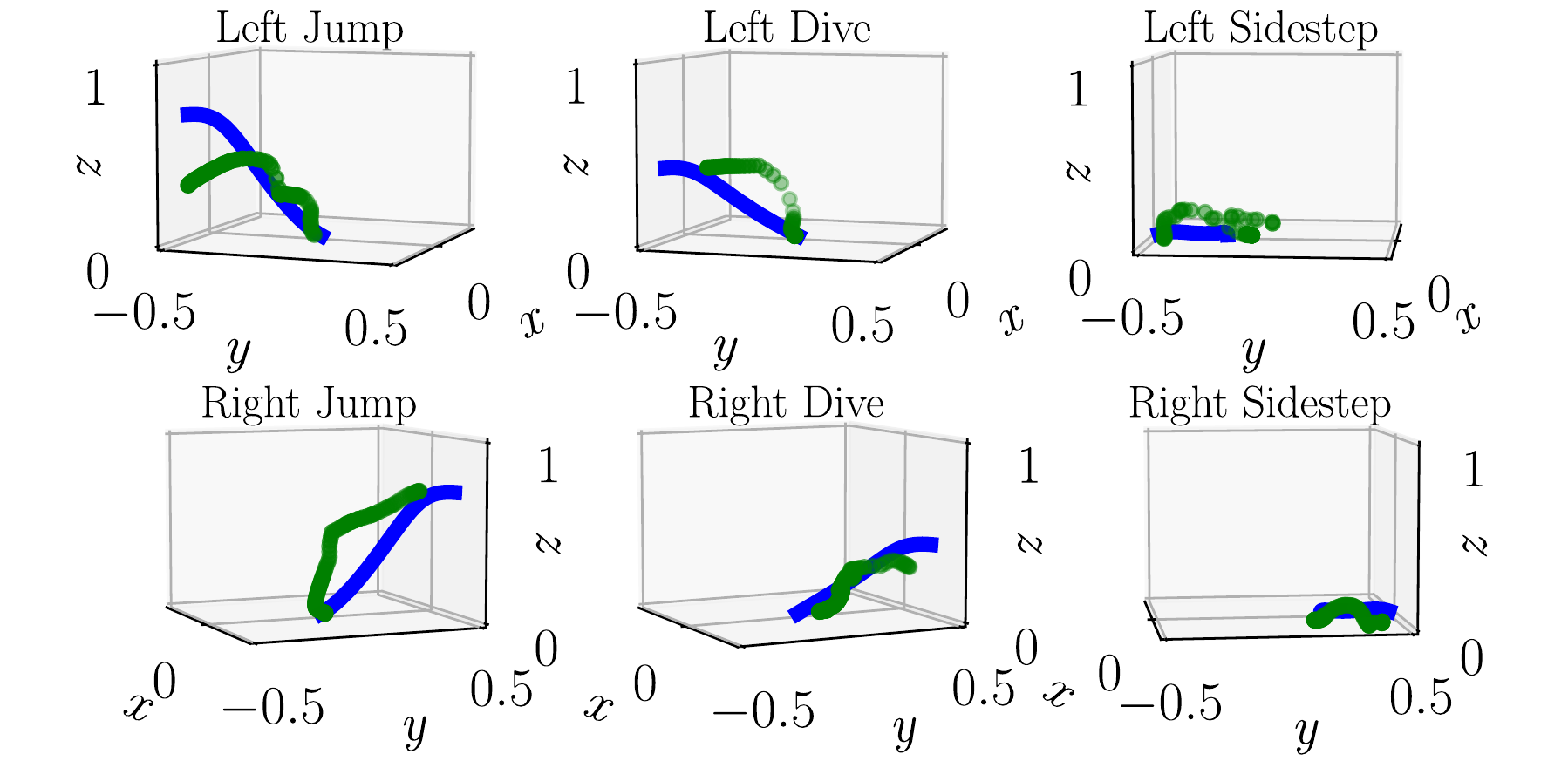}
     \end{subfigure} 
     \captionsetup{singlelinecheck = false, justification=justified}
    \caption{
     Comparision between robot's actual end-effector trajectory (green) and desired one (blue).  The actual trajectory of robot's end-effector is obtained by MoCap while the desired one is randomly specified. The average tracking error for left jumping, left diving, and left sidestepping are $0.10$, $0.15$, $0.08$ m, and those for right sides are $0.12$, $0.13$, $0.05$ m, respectively.}
     \label{fig:trackingmap3d}
     \vspace{-0.4cm}
\end{figure}

The performance of the low-level controllers on the robot hardware is firstly validated, as demonstrated in Fig.~\ref{fig:controltrack}. 
The control policies are able to produce similar maneuvers in real world (Fig.~\ref{fig:controltrack}) as in the simulation (Fig.~\ref{fig:multiskill}) without finetuning. 
Furthermore, given a random set of B\'ezier coefficients, all three policies are able to track the desired trajectories for the robot's end-effector to the best of its physical limitations, as shown in Fig.~\ref{fig:trackingmap3d}, with an average tracking error of $0.11$ m (measured by MoCap) over all trials.

\subsubsection{Goalkeeping Performance}
\begin{figure}
     \centering
     \begin{subfigure}[b]{0.15\textwidth}
         \centering
         \includegraphics[width=\textwidth]{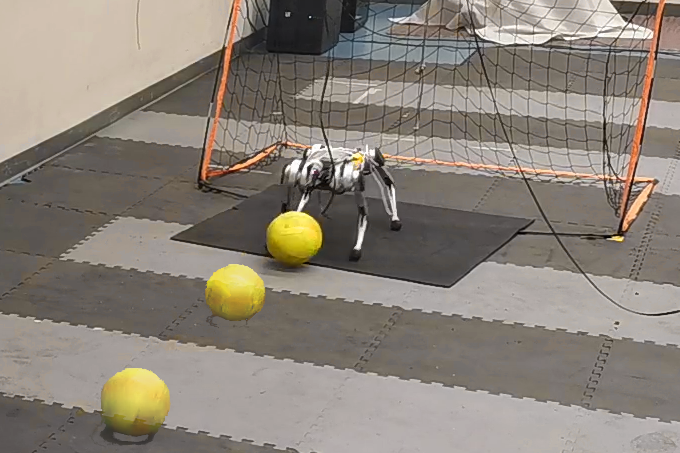}
         \setcounter{subfigure}{0}%
         \caption{Sidestep to catch nearby ground ball}
         \label{fig:expsidestep}
     \end{subfigure}
     \hfill
     \begin{subfigure}[b]{0.15\textwidth}
         \centering
         \includegraphics[width=\textwidth]{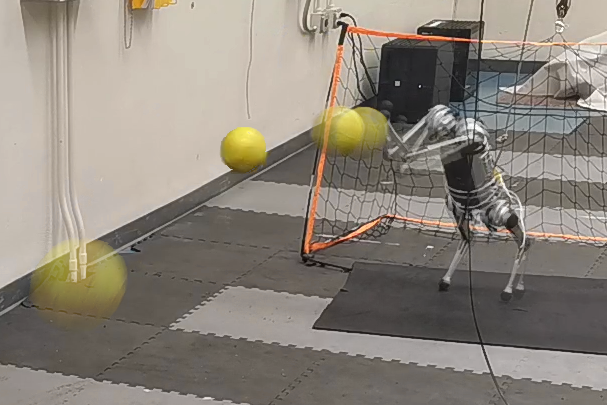}
         \setcounter{subfigure}{1}%
         \caption{Jump and punch ball away}
         \label{fig:expjump}
     \end{subfigure}
     \hfill
    \begin{subfigure}[b]{0.15\textwidth}
         \centering
         \includegraphics[width=\textwidth]{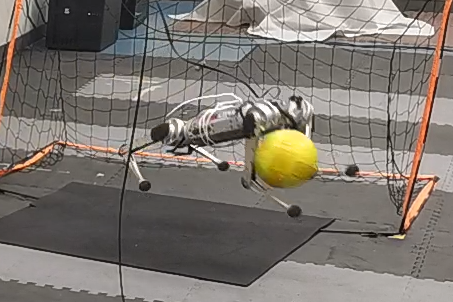}
         \setcounter{subfigure}{2}%
         \caption{Dive for mid-height ball}
         \label{fig:expdive}
     \end{subfigure}
     \newline
     \begin{subfigure}[b]{0.15\textwidth}
         \centering
         \includegraphics[width=\textwidth]{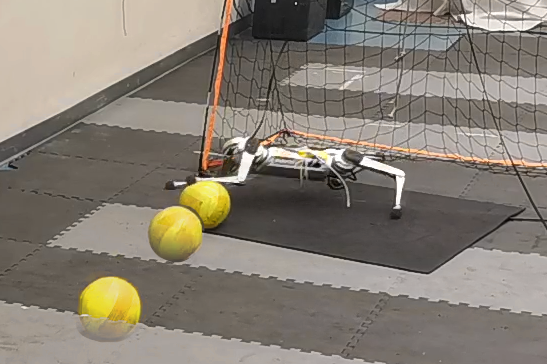}
         \setcounter{subfigure}{3}%
         \caption{Dive for corners}
         \label{fig:expdivecorner}
     \end{subfigure}
     \hfill
          \begin{subfigure}[b]{0.15\textwidth}
         \centering
         \includegraphics[width=\textwidth]{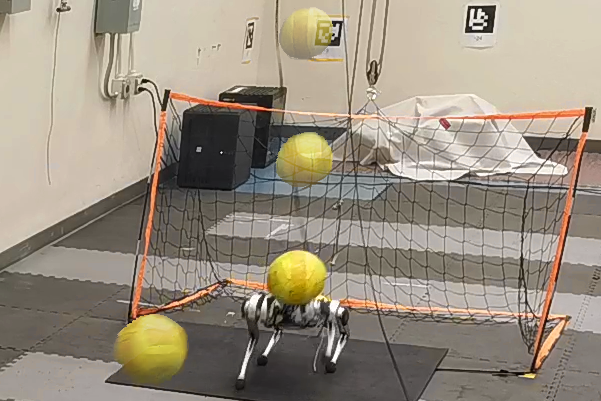}
         \setcounter{subfigure}{4}%
         \caption{Policy header skill}
         \label{fig:expheader}
     \end{subfigure}
     \hfill
     \begin{subfigure}[b]{0.15\textwidth}
         \centering
         \includegraphics[width=\textwidth]{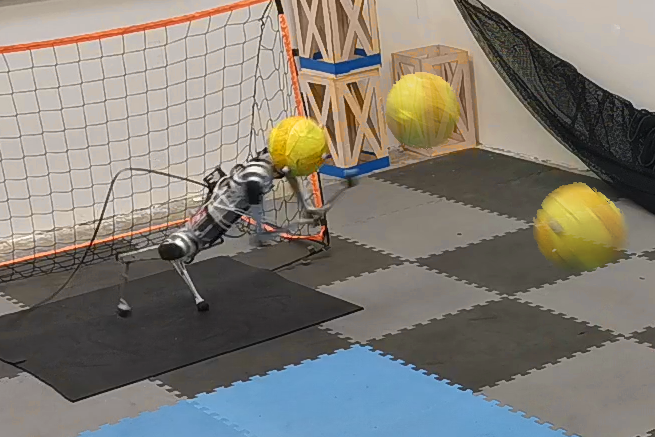}
         \setcounter{subfigure}{5}%
         \caption{Save within 0.5s}
         \label{fig:expjumpfast}
     \end{subfigure}
     \captionsetup{singlelinecheck = false, justification=justified}
        \caption{Snapshots of the real-world experiments showing the robot goalkeeper Mini Cheetah handling various scenarios. In \ref{fig:expsidestep}, the robot chooses sidestep when the ball is nearby, whereas in \ref{fig:expdivecorner} a dive save is selected as the ball is rolling towards the corner. When the ball comes high as in \ref{fig:expjump}, the robot jumps and intentionally pushes the ball away, while dive is chosen for balls in the lower half, as in \ref{fig:expdive}. Shown in \ref{fig:expheader}, the planner generalizes to leverage other parts of the body to complete the task. These experiments are conducted with a RGB-Depth camera, while the robot responded to a fast ball in less than $0.5$ second supported by MoCap in \ref{fig:expjumpfast}.}
        \label{fig:scenes3skill}
        \vspace{-0.4cm}
\end{figure}

As demonstrated in Fig.~\ref{fig:scenes3skill}, the proposed framework that utilizes three different goalkeeping skills is able to enable Mini Cheetah to save goals in different scenarios. 
For easier ones (Fig~\ref{fig:expsidestep}), the most energy-efficient way (taking a sidestep) is leveraged, while in harder cases such as in Fig~\ref{fig:expjump},\ref{fig:expjumpfast}, the robot takes a large jump and punches out the ball in the air intentionally.
In most shots, the soccer ball interception time is within $0.9$ second and the robot is able to quickly react to it.
Note that another advantage of our planner is that it may leverage the existing skills to infer other skills, such as the header in Fig.~\ref{fig:expheader}, which prevents the ball from slipping through its feet.

\begin{figure}
     \begin{subfigure}[b]{0.48\textwidth}
         \centering
         \includegraphics[width=\textwidth]{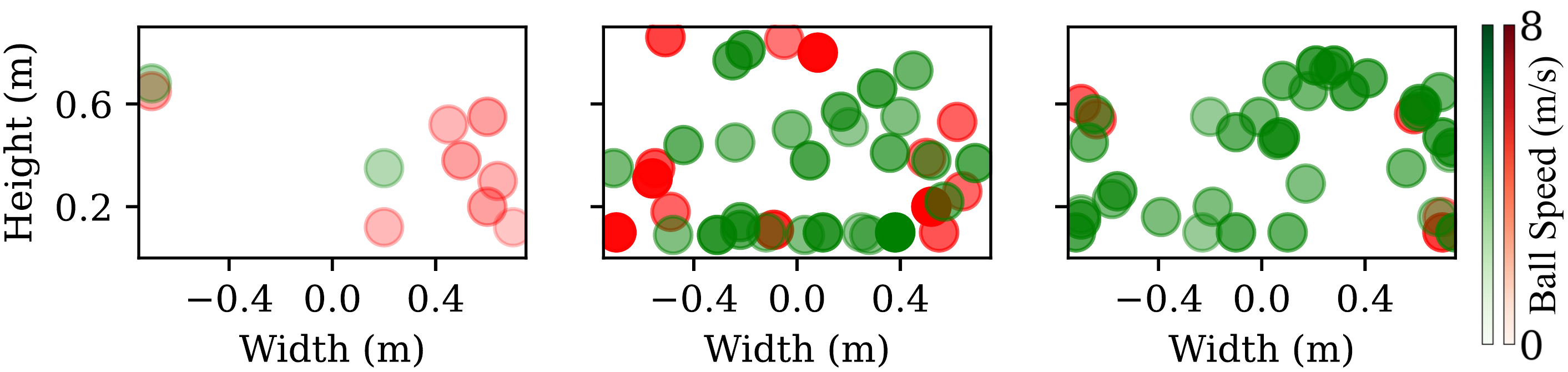}
     \end{subfigure}
     \begin{subfigure}[b]{0.15\textwidth}
         \setcounter{subfigure}{0}%
         \caption{Model-based}
         \label{fig:baselinereal}
     \end{subfigure}
     \hfill
     \begin{subfigure}[b]{0.15\textwidth}
         \centering
         \setcounter{subfigure}{1}%
            \caption{2-Skill}
         \label{fig:2skillreal}
     \end{subfigure}
     \hfill
     \begin{subfigure}[b]{0.15\textwidth}
         \centering
         \setcounter{subfigure}{2}%
         \caption{3-Skill}
         \label{fig:3skillreal}
         \end{subfigure}
     \captionsetup{singlelinecheck = false, justification=justified}
        \caption{Shot interception maps in real-world experiments. Model-based method yields a very low saving rate due to imprecise ball prediction. The lower-corner failures in \ref{fig:3skillreal} are significantly reduced compared to \ref{fig:2skillreal}, coherent with the result in simulation. The saving rates are 20.0\%, 66.7\%, and 87.5\%, respectively. These experiments are conducted using a RGB-Depth camera.}
        \label{fig:accumapreal}
    \vspace{-0.4cm}
\end{figure}

To further evaluate the performance in the goalkeeping task using the proposed framework, we conducted extensive ablation study on three methods: 1) a \textit{model-based} planner, 2) the \textit{2-skill} planner with jump and sidestep skills, and the proposed planner which utilizes~\textit{3-skill}.
The \textit{model-based} planner runs an optimization online to determine the desired B\'ezier coefficient. Like most prior work using model-based methods~\cite{gao2021optimal}, we also develop a Kalman Filter to estimate ball velocity from measurements of the ball position and dynamical model of the ball (assuming there is only gravitational force acting on the ball). 
The planner is designed to find a B\'ezier curve that can intercept the ball along the ball's predicted path and the locomotion skill is selected based on the ball height at the predicted interception point. 
The shot interception map using the \textit{model-based} planner, the \textit{2-skill} planner, and the \textit{3-skill} planner (proposed) are recorded in Fig.~\ref{fig:accumapreal}.

The \textit{model-based} planner results in the worst performance with only a $2/10$ saving rate. This is because the state estimation of the ball velocity is not reliable while the model-based planner relies on an accurate predicted interception point.
On the other hand, the learning-based planners, which are \textit{2-skill} and \textit{3-skill} planners, do not require knowing the velocity of the ball, and as a result, they demonstrate big jump in saving rates.
Both \textit{2-skill} and \textit{3-skill} planners are tested consecutively for 40 trials, and the shot interception maps are shown in Figs.~\ref{fig:2skillreal},\ref{fig:3skillreal}. 
The most frequent failure spots for the \textit{2-skill} planner occurs noticeably on the lower corners, which is innately difficult for the robot without learning the dive skill. 
In contrast, the proposed \textit{3-skill} planner ($87.5\%$ saving rate) with all three skills noticeably alleviate these corner cases and outperform the \textit{2-skill} planner by $20.9\%$ in the experiment.  

However, the limitation of the proposed framework is that the robot usually fails to save the ball whose flying time is less than $0.5$ s, considering that the entire robot motion's timespan is $0.5$ s and ball detection is delayed from the camera.

\subsubsection{Penalty Kicks with Humans and a Quadrupedal Robot}
We further showcase the capacity of the proposed goalkeeping framework by inviting human soccer players to conduct penalty shots with the robot goalkeeper. 
Penalty kicks between a quadrupedal robot soccer ball shooter developed in \cite{ji2022hierarchical} and our goalkeeper are also demonstrated. 
These experiments are recorded in the video.

\section{Conclusion and Future Work}
In conclusion, we proposed a multi-skill reinforcement learning framework that enables quadrupedal robots to function as soccer goalkeepers with precise and highly dynamic maneuvers. 
We developed a RL-based framework in simulation and demonstrated its performance with zero-shot transfer to the real world. 
The framework consists of multiple locomotion controllers specialized in specific skills (sidestep, dive, and jump) and a multi-skill manipulation planner to find the optimal skill and desired trajectory for robot's end-effector to intercept the incoming ball. 
We showcase that the multi-skill RL framework significantly outperformed a model-based planner, and was able to adequately leverage the speciality of each skill. 
In this work, we focused solely on the goalkeeping task, but the proposed framework can be extended to other scenarios, such as multi-skill soccer ball kicking. 

\section*{Acknowledgements} 
We thank Prof. S. Kim, the MIT Biomimetic Robotics Lab, and NAVER LABS for lending the Mini Cheetah. We also thank Prof. M. Mueller for use of the motion capture space and P. Kotaru and J. Chen for the help with the experiments.

{
\bibliographystyle{IEEEtran}
\bibliography{bib/bibliography}
}

\begin{acronym}
\acro{HP}{high-pass}
\acro{LP}{low-pass}
\end{acronym}

\end{document}